\pdfoutput=1
\documentclass[11pt]{article}

\usepackage[]{ACL2023}

\usepackage{times}
\usepackage{latexsym}
\usepackage{booktabs}

\usepackage[T1]{fontenc}
\usepackage[utf8]{inputenc}

\usepackage{microtype}

\usepackage{inconsolata}

\usepackage{graphicx}
\usepackage{xcolor}
\usepackage{enumitem}
\usepackage{hyperref}

\usepackage{todonotes}

\newcommand{\strict}[1]{\textsc{Strict}}
\newcommand{\strictsmall}[1]{\textsc{Strict-small}}
\newcommand{\paper}[1]{\textsc{Paper}}
\newcommand{\vision}[1]{\textsc{Multimodal}}
\newcommand{\interact}[1]{\textsc{Interaction}}

\title{BabyLM Turns 3: Call for papers for the 2025 BabyLM workshop \\
 \vspace{0.2cm} \url{https://babylm.github.io/}}

\author{
BabyLM Team 
\AND
        Lucas Charpentier \\ LTG, University of Oslo \And  
        Leshem Choshen \\ IBM Research, MIT \And  
        Ryan Cotterell \\ ETH Zürich \And  
        Mustafa Omer Gul \\ Cornell University
        \AND
        Michael Y. Hu \\ NYU \And  
        Jing Liu \\ ENS-PSL \And  
        Jaap Jumelet \\ University of Groningen \And  
        Tal Linzen \\ NYU
        \AND
        Aaron Mueller \\ Northeastern University \And  
        Candace Ross \\ Meta AI \And  
        Raj Sanjay Shah \\ Georgia Tech \And  
        Alex Warstadt \\ UC San Diego
        \AND
        Ethan Wilcox \\ Georgetown University \And  
        Adina Williams \\ Meta AI
}
\begin{document}

\maketitle

\begin{abstract}
BabyLM aims to dissolve the boundaries between cognitive modeling and language modeling. We call for both workshop papers and for researchers to join the 3rd BabyLM competition. As in previous years, we call for participants in the data-efficient pretraining challenge in the general track. This year, we also offer a new track: \interact{}. This new track encourages interactive behavior, learning from a teacher, and adapting the teaching material to the student.

We also call for papers outside the competition in any relevant areas. These include training efficiency, cognitively plausible research, weak model evaluation, and more.  

%LAST YEAR'S ABSTRACT: After last year's successful BabyLM Challenge, the competition will be hosted again in 2024/2025. The overarching goals of the challenge remain the same; however, some of the competition rules will be different. The big changes for this year's competition are as follows: First, we replace the loose track with a \emph{paper track}, which allows (for example) non-model-based submissions, novel cognitively-inspired benchmarks, or analysis techniques. Second, we are relaxing the rules around pretraining data, and will now \emph{allow participants to construct their own datasets} provided they stay within the 100M-word or 10M-word budget. Third, we introduce a \emph{multimodal vision-and-language track}, and will release a corpus of 50\% text-only and 50\% image--text multimodal data as a starting point for LM model training. The purpose of this CfP is to provide rules for this year's challenge, explain these rule changes and their rationale in greater detail, give a timeline of this year's competition, and provide answers to frequently asked questions from last year's challenge.
% ADD: Please see website listed above for date and rules updates.

\end{abstract}

\section{Introduction: BabyLM %[Ethan, Alex]
    }

The goals of BabyLM are to bring together multiple disciplines to answer an enduring question: how can a computational system learn language from limited inputs? Cognitive scientists investigate this question by trying to understand how \emph{humans} learn their native language during childhood. Computer scientists tackle this question by attempting to build efficient machine-learning systems to accomplish this task. BabyLM brings these two communities together, asking how insights from cognitive science can be used to assemble more sample-efficient language models and how language modeling architectures can inspire research in cognitive science.

Previously, BabyLM has been organized as a competition, challenging participants to train a language model on a human-sized amount of data, up to 100 million words. This year, we expand the scope of BabyLM by presenting it as a \emph{workshop}. While we will still run the competition, we also invite original research papers at the intersection of cognitive science and language modeling without entry into any competition track (See suggested topics in \S\ref{sec:topics}).

% LAST Year's Introduction

%After last year's successful BabyLM Challenge, the competition will be hosted again in 2024/2025. The overarching goals of the challenge remain the same, however, some of the competition rules will be different. The purpose of this CFP is to provide a timeline of this year's competition, explain the rule changes and their rationale, and provide answers to frequently asked questions from last year's challenge.

%As before, the goal of the BabyLM shared task is to incentivize researchers interested in pretraining and/or cognitive modeling to focus their efforts on optimizing pretraining given data limitations inspired by human development. Additionally, we hope to democratize research on pertaining by drawing attention to open problems that can be addressed on a university budget. Please see the previous call for paper \citep{warstadt2023call} or the introduction to last year's proceedings \citep{warstadt2023findings} for a deeper discussion of the rationale behind the challenge.

In terms of the BabyLM challenge, this year's iteration will remain largely the same but with a few key differences, which we list below:

\begin{itemize}[leftmargin=0.5cm,rightmargin=0.5cm]

    \item We are debuting a new \interact{} \emph{track}, exploring how feedback and interaction can assist with sample-efficient language modeling. This track will allow pre-trained language models to serve as teacher models; however, student models are still required to train on 100 million words or less. 

    \item We offer a \strict{} and \strictsmall{} track, challenging participants to train on 100M and 10M words, respectively. We will also re-release the \emph{\vision{} track} from last year's challenge. Although we had limited participants in this track in the previous iteration of the challenge, we hope that by re-releasing the track, we can encourage more participation in multi-modal language modeling research.
    
    \item This year, we will impose additional \emph{compute limitations} on all challenge tracks. Models may not conduct more than 10 epochs over their training data. The motivation behind this change and details on how it is instantiated for all tracks are detailed further in Section \ref{sec:training-requirements}.
    
\end{itemize}

\section{Key Dates}

\emph{Tentative Timeline}: We will accept submissions through ACL Rolling Review (ARR) or directly through OpenReview. Paper submissions to the workshop can ignore competition entry deadlines. Our anticipated timeline. Dates will be determined based on eventual ARR deadlines.

\begin{itemize}[leftmargin=0.5cm,rightmargin=0.5cm,itemsep=0em]
    \item \textbf{Early February:} CfP released
    \item \textbf{End of February:} Training data released
    \item \textbf{End of April:} Evaluation pipeline released
    \item \textbf{May 19:} ARR submission deadline
    \item \textbf{Mid July:} Direct submissions deadline
    \item \textbf{Mid August:} Direct submission reviews due, ARR commitment deadline
    \item \textbf{Early September:} Decisions released
    \item \textbf{Mid September:} Camera ready due
    \item \textbf{5--9 Nov:} Workshop @ EMNLP in Suzhou
\end{itemize}

\section{Paper guidelines and submission% [Michael, Leshem]
}
\subsection{Topics} \label{sec:topics}
The BabyLM workshop encourages interdisciplinary submissions at the interface of language modeling, cognitive science, language acquisition, and/or evaluation. To this end, we will accept papers on a variety of topics, including but not limited to the following:
\begin{itemize}[noitemsep]
    \item Data-efficient architectures and training techniques
    \item Data curation for efficient training
    \item Cognitively and linguistically inspired language modeling and evaluation
    \item Scaling laws; large and small model comparisons
    \item Cognitively inspired multimodal modeling or evaluation
\end{itemize}

% \subsection{Formatting}
% The paper should follow the EMNLP format. This includes length and anonymity requirements upon submission. Reviewing will be double-blind.

\subsection{Paper submission} Submissions will be made through OpenReview. Submissions can be full archival papers or non-archival upon request and can be up to eight pages in length. Formatting requirements will follow standards for EMNLP 2025 workshops. This includes length and anonymity requirements upon submission. Reviewing will be double-blind. As before, we do allow dual submission; however, we do not allow dual publication.

% LAST YEAR'S TEXT

%Submissions will be made through OpenReview. All submissions will be full archival papers and can be up to eight pages in length. Formatting requirements for submissions will be announced at a later date, once the presentation venue has been finalized. As before, we do allow dual-submission, however, we do not allow dual publication.

\subsection{Review \& Publication}
BabyLM will hold a review process, as happens in most workshops. Papers submitted to the workshop will be evaluated on merit and relevance. For competition participants, acceptance is lenient. We plan only to reject competition submissions that make incorrect or unjustified claims, that have significant technical issues, that do not reveal enough methodological details for replication, or that demonstrate only minimal time investment. Feedback will largely be directed toward improving submissions.

% LAST YEAR'S TEXT:

%BabyLM will hold its own review process. The acceptance criteria are based on soundness: We plan only to reject submissions that make incorrect or unjustified claims. Other feedback will be directed at the improvement of submissions.

\section{Competition Details}

\subsection{Track Rules}

The second BabyLM Challenge includes four competition tracks: \textbf{\strict}, \textbf{\strictsmall}, and \textbf{\vision} and \textbf{\interact{}}. 
\textbf{Note that participation in one of the competition tracks is not a prerequisite for submitting to the workshop.}

\paragraph{New track: Interactivity.} The \textbf{\interact{}} track debuts this year to allow for interaction between multiple agents during training.
We will distinguish between a \textbf{submission model}, i.e., the participants' entry into the competition, and an \textbf{external model}, i.e., a secondary model used in the training pipeline of the submission model but not submitted to the competition.
External models must come from a predetermined list of models available on the BabyLM website. External models may be fine-tuned or distilled without restriction.
However, the submission model must be exposed to no more than 100M word tokens (multiple exposures allowed, e.g., epochs); this word count includes text generated by external models \emph{and} pre-existing corpora. 
Additionally, the submission model may not generate more than 100M words during the training process.
Finally, the external model's weights, hidden states, or output distribution cannot be revealed to the submission model.

\paragraph{Continuing Tracks}
The rules for \textbf{\strict}, \textbf{\strictsmall}, and \textbf{\vision} remain unchanged from last year's competition \citep{choshen2024call}.
We quote:
\begin{quote}
    The \strict{} and \strictsmall{} tracks require that submissions be trained on a corpus of 100M words or less (in \strict{}) and 10M words or less (in \strictsmall{}). These tracks do \textit{not} require that participants use the official BabyLM corpus, although we will still provide an updated version of this dataset for participants interested in using it. Models in these tracks will be evaluated on language-only evaluation tasks.

    In the \vision{} track, participants will train multi-modal image-text models. Participants can use whatever training procedure they wish, as long as models can provide (pseudo) log-likelihoods to strings of text, conditioned on an image. Participants are free to use whatever data they wish, as long as the dataset is within a 100M word budget. To facilitate easier participation in this track, we will re-release a suggested multimodal dataset that consists of 50\% text-only and 50\% paired image--text data. Submissions to this track will be evaluated on language-only tasks, as well as multi-modal tasks.
\end{quote}

\subsection{Training Requirements}\label{sec:training-requirements}
% \paragraph{Intermediate Checkpoints}

%\textbf{TODO} -- update this language to being a suggestion for epoch limitation; save a checkpoint after  every 1M words up to 10M, then 10M words up to 100M for a total of N checkpoints

% New this year, for deeper analysis, we will require submitting intermediate checkpoints. These checkpoints should be saved (i) every 1M words up to 10M words (ii) 10M words up to 100M words and (iii) every 100M words thereafter.

% \paragraph{Epoch Limitation}
New this year, we introduce two additional requirements for your submitted models:
\begin{enumerate}
    \item Models submitted to the leaderboard can now only be exposed to a fixed amount of input.
    \item Intermediate model checkpoints must be submitted as well to test for learning speed and model behavior dynamics.
\end{enumerate}
Below, we explain these choices in more detail.

\paragraph{Training Duration Limitations}
As of this year, we require the submission of results for models trained on a fixed amount of input (\emph{counting} repeated exposures), in particular \textbf{after at most 100M words for the \strictsmall{} track and after at most 1B words for all other tracks}.
In most cases, this will mean after 10 epochs on the standard BabyLM corpora. However, because what counts as an epoch may differ across submissions, we instead quantify training by the number of whitespace-separated input words.
While participants are welcome to train for longer and report this in their paper, we will only include the model checkpoints that follow these limitations in the leaderboard.
Note that it is also allowed to submit a model that is trained on less data: the 100M and 1B word limits are \textit{an upper bound} on data exposure. For the \interact{} track, the number of words seen by the submission model is considered to be the sum of the number of input words and generated tokens.

%the number of words seen by the submission model is considered to be the number of input words from external sources, such as the BabyLM corpus or teacher generations, and excludes the student's own outputs. If a challenge submission incorporates interaction only after an initial pre-training stage (that is, if a student model is first trained up to the limit of 1B words and is then further finetuned via RL using only its own generated outputs), then checkpoints before and after the interaction process should be provided. In this case, the final model checkpoint following interactive training would be the leaderboard entry.

% Note that this is a minimal requirement: Results from additional checkpoints are also encouraged, in which case we recommend evaluating at exponentially growing intervals (e.g., 100K, 316K, 1M, ...).
% We will also allow entrants to submit only a single checkpoint at 100M words (or 10M words for \strictsmall), or a checkpoint at 100M (or 10M) words and a second checkpoint at less than 1B (or 100M) words.
% Results from models trained on more than 1B words can be submitted but will not count towards competition performance.

\paragraph{Intermediate Checkpoints}
We also require the submission of intermediate model checkpoints to the HuggingFace Hub.
These checkpoints will be used for the updated evaluation pipeline (\S\ref{sec:eval}), to measure aspects related to learning efficiency and language acquisition.
The checkpoints we require will be at increasing intervals: every 1M words until 10M words are seen, every 10M words until 100M words are seen, and (for the tracks other than \strictsmall{}) every 100M words until 1B words are seen.
More precise details about the evaluation of these intermediate checkpoints will be announced with the release of the evaluation pipeline.

\paragraph{Motivation}
In previous years, we provided no such requirements. 
One motivation for doing so this year is that the training dynamics of LMs can be compared to the learning trajectories of children, which is valuable from a cognitive modeling perspective.
Furthermore, one of the conclusions of the 2024 BabyLM Challenge is that more compute is correlated with higher performance.
This runs counter to the goals of BabyLM in several ways:
First, one goal of BabyLM is developmentally plausible training, but children do not experience repeated exposure to their input.
While we allow that memories of inputs could have an impact on learning beyond the initial exposure, we judge 10s or 100s of repeated exposures to every input to be developmentally implausible.
Second, another goal of BabyLM is to democratize pretraining research, but large numbers of training epochs require greater computational resources that are not available to all participants.
As a consequence, well-funded or well-equipped research groups have a significant advantage if no limitation is applied.
This advantage does not disappear with this restriction, as well-funded groups may be able to afford more hyperparameter searches and prototyping, but these efforts will at least lead to training recipes that can be reproduced in future cycles by less well-funded groups.

We have \emph{not} chosen to restrict the amount of compute.
While such a restriction might be ideal from the perspective of democratization, it is less clear (but by some estimates unlikely) that BabyLM submissions exceed the computation available to children \citep{sandberg2008whole}.
Furthermore, a requirement to compute FLOPs is more technically demanding than one to count the amount of data seen, and it could deter participation with limited additional advantages.

Finally, this restriction only applies to competition entries.
Workshop papers are not required to include models with 10 or fewer training epochs, though this is, of course, encouraged.

\subsection{Provided Dataset}\label{sec:dataset}

\begin{table*}[t]
    \centering
    \resizebox{\linewidth}{!}{
    \begin{tabular}{llrrr}
    \toprule
    Dataset & Description & \# Words (multimodal track) & \# Words (strict track) & \# Images\\
    \midrule
    Localized Narratives \cite{LocalizedNarratives} & Image Caption & 27M & -- & 0.6M \\
    Conceptual Captions 3M \cite{CC3M} & Image Caption & 23M & -- & 2.3M \\
    CHILDES \citep{macwhinney2000childes} & Child-directed speech & 15M & 29M & -- \\
    British National Corpus (BNC), dialogue portion & Dialogue & 4M & 8M & -- \\
    Project Gutenberg (children's stories) \citep{gerlach-2018-gutenberg} & Written English & 13M & 26M & -- \\
    OpenSubtitles \citep{lison-tiedemann-2016-opensubtitles2016} & Movie subtitles & 10M & 20M & -- \\
    Simple English Wikipedia & Written Simple English & 7M & 15M & -- \\
    Switchboard Dialog Act Corpus \citep{Stolcke-etal:2000} & Dialogue & $<$1M & 1M & -- \\
    \midrule
    \emph{Total} & -- & 100M & 100M & 2.9M\\
    \bottomrule
    \end{tabular}}
    \caption{Datasets for the multimodal and strict tracks of the BabyLM competition. The data has not changed from the 2nd BabyLM Challenge \citep{choshen2024call} Word counts are approximate and subject to slight changes. 
    % \textsuperscript{1}\url{https://google.github.io/localized-narratives/}\ \ \ \textsuperscript{2}\url{https://ai.google.com/research/ConceptualCaptions/download}
    \looseness=-1}
    \label{tab:data}
\end{table*}

We provide the same training datasets as in last year's competition \citep{choshen2024call} at this \href{https://osf.io/ad7qg/}{link}; see Table~\ref{tab:data} for dataset composition statistics.
This includes:
\begin{itemize}[noitemsep]
    \item 100M word \strict{} dataset
    \item 10M word \strictsmall{} dataset
    \item 100M word + image \vision{} dataset.
    
\end{itemize}

The previous BabyLM versions portray a creative use of diverse data sources and modalities for language model development. We will provide a list of downloadable datasets that previous \emph{submissions} to BabyLM used to train models.
As with this year, last year's competition allowed participants to train on an original dataset of 100M words or less (or 10M words). 
Those datasets have been made publicly available at the following \href{https://docs.google.com/spreadsheets/d/1R4spgWHdSkYDZceaXHOdj0c-wMTSrn7ny7kC1lOf0ko/edit?usp=sharing}{link}. %\todo{Verify if the link looks good}

\subsection{Evaluation}\label{sec:eval}

% PREVIOUS YEAR'S TEXT:

As in previous years, we will distribute an open-source evaluation pipeline. This year, the evaluation pipeline will be written from scratch so as to make the structure of the repository significantly simpler than in previous years. This will make it easier for participants to adapt it to their needs or unique architectures and debug any potential issues. We will have a HuggingFace implemented version as well as one using PyTorch modules to allow testing without having to create a HuggingFace implementation of the model. % Lucas, Aaron
Much of the evaluation will be based on zero-shot probability comparisons of two text sequences, as in previous years. 

This year, we will additionally include tasks that measure \textit{psychometric fit} to human language learners. % Jaap
The tasks that we will add to the evaluation suite will focus on two aspects of a model being `human-like': i) connecting model behavior and internals to cognitive aspects of human language processing, such as reading time prediction, and ii) assessing how human-like a model's generalizations are on various tasks related to reasoning and morphology. Human-likeness metrics will be considered separate from accuracy metrics, such that a system could win either with respect to NLP task performance \emph{or} human-likeness. We plan to give separate awards for both metrics.

As in previous years, we will be releasing hidden evaluations to control for overfitting to the public evaluation tasks. These will be released no less than two weeks before the model submission deadline.

More details about the evaluation pipeline and the set of tasks will be released subsequently.

\subsection{Baselines}

We will release a series of baseline models. Similar to the previous year, we will release baselines based on the winning submissions from the last year. For the \strict{} and \strictsmall{} tracks, we will release the following baselines: GPT-BERT~\citep{charpentier2024gptbertboth}, the winning submission from the last year, and GPT-2 Small~\citep{radford2019language}, as a purely autoregressive baseline. For the \emph{\vision{}} track, we will be re-releasing the GIT~\citep{wang2022git} and Flamingo~\citep{alayrac2022flamingo} baselines, as no submissions were able to outperform them last year.

For the \interact{} track, we will provide two baselines intended to act as examples of how feedback and interaction can be instantiated during language model training. 

We provide a baseline that explores how communicative feedback can be integrated with language model training \citep{nikolaus2021modeling,ma2024babysit}. We train a reward model on child-caregiver interactions to predict when a child’s utterance triggers a communicative response (hereafter CR) based on prior research \citep{nikolaus2022communicative}. In other words, for a given utterance, the reward model predicts whether it would likely be followed by a CR by the caregiver. The goal of using a reward model is to provide parent-like rewards to the input-based baseline’s own produced utterances in the fine-tuning stage. We train the binary reward model based on child-caregiver conversations by fine-tuning \textit{deberta-v3-xsmall} \cite{he2021debertav3} as follows: If an utterance produced by the target language model followed by a CR, it is assigned a reward value of 0, and 1 otherwise. We fine-tune the language models pre-trained on the BabyLM corpus using Proximal Policy Optimization (PPO). For each fine-tuning step, we a) sample utterances from the target language model \citep{choshen2024call}(with temperature sampling in which temperature is set as default = 1.0), b) compute the corresponding rewards from the reward model, and c) update the language model’s weights using PPO. The best checkpoint is selected based on the mean reward. We use rejection sampling to discourage too-long and too-short utterances: setting as -1 for all generated utterances less than 3 tokens long or without the end-of-sequence token within 20 tokens.

We also provide a baseline that explores how corrections in natural language can be incorporated into language model training. We split training into 20 rounds of interaction. At each round, the student model, chosen to be GPT-2 Small~\citep{radford2019language}, is given incomplete data points sampled from the BabyLM training corpus. For each data point, the student samples a completion. The teacher model, chosen to be Llama-3.1 Instruct 8B~\citep{dubey2024llama}, is then prompted to revise the student completion based on grammaticality, coherence, and relevance to the input. The student model is then first trained with the language modeling loss on the full teacher-corrected datapoint and is then further finetuned with SimPO~\citep{meng2024simpo}, a preference optimization algorithm, wherein the teacher and student completions are the winning and losing responses respectively.

These baselines are meant to encourage participants to innovate and improve beyond existing models and approaches. 

% PREVIOUS YEAR'S TEXT:

%We will release a series of baseline models. As opposed to last year's baselines, which were trained relatively naively, this year's baseline will be based on the winning submissions from last year. For the \strict{} and \strictsmall{} tracks, we will release the following baselines:  GPT2 (decoder-only; \citealp{radford2019language}), LTG-Bert (encoder-only; \citealp{LTGBert}), and Contextualizer (decoder-only; \citealp{Contextualizer}). 
%For the \vision{} track, we will release the GIT~\cite{wang2022git} and Flamingo~\cite{alayrac2022flamingo} baselines.
%These baselines are meant to encourage participants to innovate and improve beyond existing models.

\subsection{Competition Submission %[Aaron, Lucas, Raj]
}

%As with last year, submissions will be made using the Dynabench platform, which is an online platform for dynamic data collection and model benchmarking.\footnote{\url{https://dynabench.org/}}

Competition paper submissions will be made through OpenReview. This will include links to models and predictions, as well as links to custom datasets if applicable.

Predictions can also optionally be uploaded to a HuggingFace leaderboard at any time (including after the deadline); the leaderboard's contents will be made public after paper acceptance notifications. It will be possible to submit to the leaderboard after the deadline; this will allow future innovations to build on top of the efforts of competition participants.

\vspace{0.2cm}
\noindent\textbf{What you need to submit:}
\vspace{-0.3cm}
\begin{itemize}[leftmargin=0.5cm,rightmargin=0.5cm, itemsep=0cm]
    \item A link where we can download the model (any file-hosting service will do).
    \item Model predictions in a format compatible with the evaluation pipeline.
    \item A datasheet describing the composition of the custom dataset and containing a download link (if not using a BabyLM-provided corpus)
    \item If submitting to the \interact{} track, fine-tuning, and distillation data for the external model (if any), and any data generated by the submission or external model
\end{itemize}
\vspace{-0.3cm}

\section{FAQs %[Ethan, Leshem, Michael]
}
\paragraph{Can I do cool idea X?}
If it is interesting, innovative, or may result in important findings, we want you to try it! If you think the rules are holding you back from submitting to the competition, please reach out to the organizers. In the worst (or best) case scenario, it can be an interesting workshop paper.

\paragraph{Why doesn't BabyLM do cool idea X?}
Maybe we haven't thought about it; please reach out. We value proactivity.

\paragraph{Can papers be submitted to multiple tracks?} 
Yes. For example, a single paper can describe models that are submitted separately to the \strict{} and \interact{} tracks. 

\paragraph{Can I submit a paper about my work?}
Yes, we require that \emph{all} competition submissions be accompanied by a paper, which can be up to eight pages in length (though it does not need to be). Papers will be published in an archival format. All papers can describe experiments and analyses beyond the scope of the competition.

\paragraph{Can I submit additional evaluation metrics?}
Yes, you may submit additional evaluation metrics alongside a competition model in the \strict{}, \strictsmall{}, and \interact{} tracks. This type of contribution is especially encouraged for workshop submissions.

Moreover, we accept analysis and insightful findings on previous submissions or related topics and especially welcome evaluation that works well for small models but evaluates meaningful aspects. If you believe you know of an evaluation that we should use throughout the competition, please contact us.

\paragraph{What training regimes are permitted?}
Any training objective/regime is permitted as long as the data restrictions are followed. If you use ancillary models, for example, in the case of reranking or data augmentation, the training data for these models is counted towards your 100M word budget. This applies to all tracks, including \interact{} track; so, for example, while you can use the external model to produce POS tags, you cannot use an off-the-shelf POS tagger in your pipeline.

For evaluation purposes, we require that the model provides a function to score a sequence of words without the need for additional fine-tuning.

\paragraph{Are there any limits on hyperparameters?}
No. But please share at the end what you found so we can learn from your efforts.

\paragraph{Are there any limits on the number of epochs?}
This year, yes. Refer to the ``Training Duration Limitation'' paragraph of Section \ref{sec:training-requirements} for more details.

%Models in the \strict{} and \strictsmall{} can train for only 10 epochs, maximum. We do not impose this limit for models in the \vision{} or \interact{} tracks.
%No. We put no restrictions on the number of epochs, for several reasons: First, from an engineering perspective, training LMs with SGD tends to require multiple epochs at these scales to achieve peak performance. Second, from a cognitive perspective, humans have a memory of linguistic experience and can continue to access and learn from these memories. Third, we try not to make a stand on implementations to allow the most freedom for innovation. Our internal results suggest, however, that under regular circumstances, over-training on more than a couple epochs give minor gains at most. 

\paragraph{Can I use external tools?}
Yes, but if they are learned on language, their tokens are counted towards the 100M. That means one can train on the same text, both a tokenizer, a parser, an LM, etc., or on parts of the 100M, but the sum of all text seen by all training can not surpass the amount of text allowed. This raises the question of synthetic data, which is allowed under some restrictions. You may generate the 100M tokens in any legal way you like (yes, distilling or writing your own is fair, if you figure out what text facilitates learning, it is interesting regardless of how to gather such text), you may also train eventually on more than 100M words by augmentation, however, that only works in a closed system, i.e., the augmenters' training data counts toward the limit, so, for example, training two LMs on half of the words, and then having them generate more words and training a model on both the original data and the new one is legit (and it was not tested in the previous competition, so even the example itself is interesting). \\
Note that the \interact{} track has an additional tool allowed (the world to interact with).

\paragraph{I have different modalities that can help}
If it is not linguistic data, prove it, last year’s submissions did not gain from non-linguistic grounding, but we encourage such scientific questions. If it is linguistic in nature (e.g., audio), then the words should still count towards the overall number of learned words.

\subsection{\interact{}}
\paragraph{Can I get non-verbal cues from the teacher?}
Yes. Note, however, that the student’s outputs are limited.

\section{Organizing Committee}
(Alphabetical by last name) Lucas Charpentier, Leshem Choshen, Ryan Cotterell, Mustafa Omer Gul, Michael Hu, Jaap Jumelet, Tal Linzen, Jing Liu, Aaron Mueller, Candace Ross, Raj Sanjay Shah, Alex Warstadt, Ethan Wilcox, and Adina Williams. Feel free to contact members of the organizing committee at: \texttt{leshem.choshen@mail.huji.ac.il}, \texttt{aa.mueller@northeastern.edu}, \texttt{alexwarstadt@gmail.com}

\bibliography{custom}
\bibliographystyle{acl_natbib}

\appendix

\section{Baseline Implementation Details}

\subsection{RLHF Details}
To obtain a diverse set of produced utterances, we prompt the model with short beginnings of utterances from the language modeling training data (1 to 2 tokens). Additionally, we add an entropy regularization term (0.001) to the loss. Further, to counteract language drift, which can be an issue in RL fine-tuning studies, we added a small language modeling loss regularization term (weighted by 0.001) to the loss. We set the target KL divergence to 2. All other PPO hyperparameters were not changed from the default values implemented in the Huggingface TRL library \footnote{https://huggingface.co/docs/trl/en/index}.

\subsection{Preference Optimization Baseline Details}

\paragraph{Dataset Construction} Prior to training, we split each constituent dataset of the BabyLM corpus into 20 equally sized chunks. At each round, a chunk is sampled at random from each constituent dataset without replacement. Each chunk is then split into data points consisting of 512 tokens. 

The student is provided the first 256 tokens of each data point as context for generation. We then sample student completions with nucleus sampling~\citep{Holtzman2020The} where $p=0.8$. Teacher corrections are similarly sampled using nucleus sampling with $p=0.8$, using the prompt shown in Figure \ref{fig:teacher_prompt}.

\paragraph{Training Hyperparameters} We optimize the student model with AdamW~\citep{loshchilov2018decoupled} with a learning rate of 0.00005 and set $\beta=2$ and $\gamma=1$ for SimPO. We add the language modeling loss on the winning completion, with a scaling coefficient of 0.2, as a regularizer during preference optimization training, following \citet{dubey2024llama}.

At the start of each round, model parameters are initialized from the best-performing checkpoint of the previous round, determined by perplexity on the BabyLM validation set.

\begin{figure*}[t!]
    \centering
    \fbox{
        \parbox{\dimexpr\textwidth-2\fboxsep-2\fboxrule\relax}{\small
        \textbf{Correction Prompt}: \\
\small [User] You will be given a partial text (labeled ``Partial Text'') and a completion of said text produced by a student of English (labeled ``Student Completion''). Your goal is to produce a corrected version of the student's completion. This corrected version should be grammatically correct, coherent and relevant to the initial partial text. If the student's response is incomprehensible, output your own independent completion. You should only provide your own completion without any added commentary or feedback. \\

Partial Text: <student input> \\
Student Completion: <student completion> \\

Now produce your own completion of the Partial Text. Do not include any external commentary. \\

[Assistant] Partial Text: <student input> \\
Corrected Completion: 
        }
    }

    \caption{The prompt given to the teacher model to sample corrected versions of the student's completions.}
    \label{fig:teacher_prompt}
\end{figure*}

\end{document}